\documentclass[letterpaper, 10 pt, conference]{ieeeconf}  

\IEEEoverridecommandlockouts                              

\usepackage{graphicx}
\usepackage{amsmath}
\usepackage{booktabs}
\usepackage{multirow} 
\usepackage{hyperref}
\usepackage{adjustbox}
\usepackage{subcaption}  
\usepackage{float}       
\usepackage{threeparttable} 

\title{\LARGE \bf
WinTA-GIL: Windowed Trajectory Alignment for GNSS-IMU-LiDAR Heading Refinement in Intermittent Signal Environments
}

\author{Kaixin Feng$^{1}$, Zhichao Wen$^{1}$, Zhaohong Liao$^{2}$, Xin Xia$^{3}$, You Li$^{1}$%
\thanks{*This work was partly supported by the National Natural Science Foundation of China (42274052) and the Wuhan Natural Science Foundation Project (2025041001010363).}%
\thanks{$^{1}$Kaixin Feng, Zhichao Wen, and You Li are with the State Key Laboratory of Information Engineering in Surveying, Mapping and Remote Sensing (LIESMARS), Wuhan University, Wuhan, 430072, China. You Li is the corresponding author. {\tt\small \href{mailto:kaixinfeng@whu.edu.cn}{kaixinfeng@whu.edu.cn}}, {\tt\small \href{mailto:zhichaowen@whu.edu.cn}{zhichaowen@whu.edu.cn}}, {\tt\small \href{mailto:liyou@whu.edu.cn}{liyou@whu.edu.cn}}}%
\thanks{$^{2}$Zhaohong Liao is with the School of Remote Sensing and Information Engineering, Wuhan University, Wuhan, China. {\tt\small \href{mailto:liaozhaohong@whu.edu.cn}{liaozhaohong@whu.edu.cn}}}%
\thanks{$^{3}$Xin Xia is with the College of Engineering and Computer Science, University of Michigan-Dearborn, Dearborn, MI 48128, USA. {\tt\small \href{mailto:xinxia@umich.edu}{xinxia@umich.edu}}}%
}

\begin{document}

\maketitle
\thispagestyle{empty}
\pagestyle{empty}

\begin{abstract}
Although multi-source fusion positioning systems have achieved significant progress, accurate and reliable heading estimation remains a critical challenge due to the lack of gravitational constraints and the inherent weak observability of heading in complex environments. Most existing methodologies are specifically tailored for the startup phase, relying on a singular initial alignment to establish the heading reference. Consequently, these approaches lack the adaptability required to refine heading estimates dynamically, which renders the system highly vulnerable to accumulated drift and observation noise during prolonged navigation or immediately following GNSS signal outages. To address these limitations, this paper proposes WinTA-GIL, a novel heading refinement framework that integrates information from Global Navigation Satellite System (GNSS), Inertial Measurement Unit (IMU), and Light Detection and Ranging (LiDAR) through a temporal window-based optimization strategy. Unlike conventional alignment methods restricted to the startup phase, WinTA-GIL leverages high-precision local trajectories from LiDAR-Inertial Odometry (LIO) to register against filtered GNSS observations. This approach transforms heading estimation into a repeatable, trajectory-based consistency optimization problem. In particular, an adaptive re-estimation mechanism based on state discrimination is incorporated to trigger heading corrections whenever necessary, thereby effectively suppressing the inertial drift accumulated during challenging conditions. Extensive experiments on both open-source and self-collected datasets demonstrate that WinTA-GIL significantly outperforms state-of-the-art approaches in both estimation accuracy and system robustness.
\end{abstract}

\section{INTRODUCTION}

\begin{figure}[!t]
    \centering
    \includegraphics[width=0.45\textwidth]{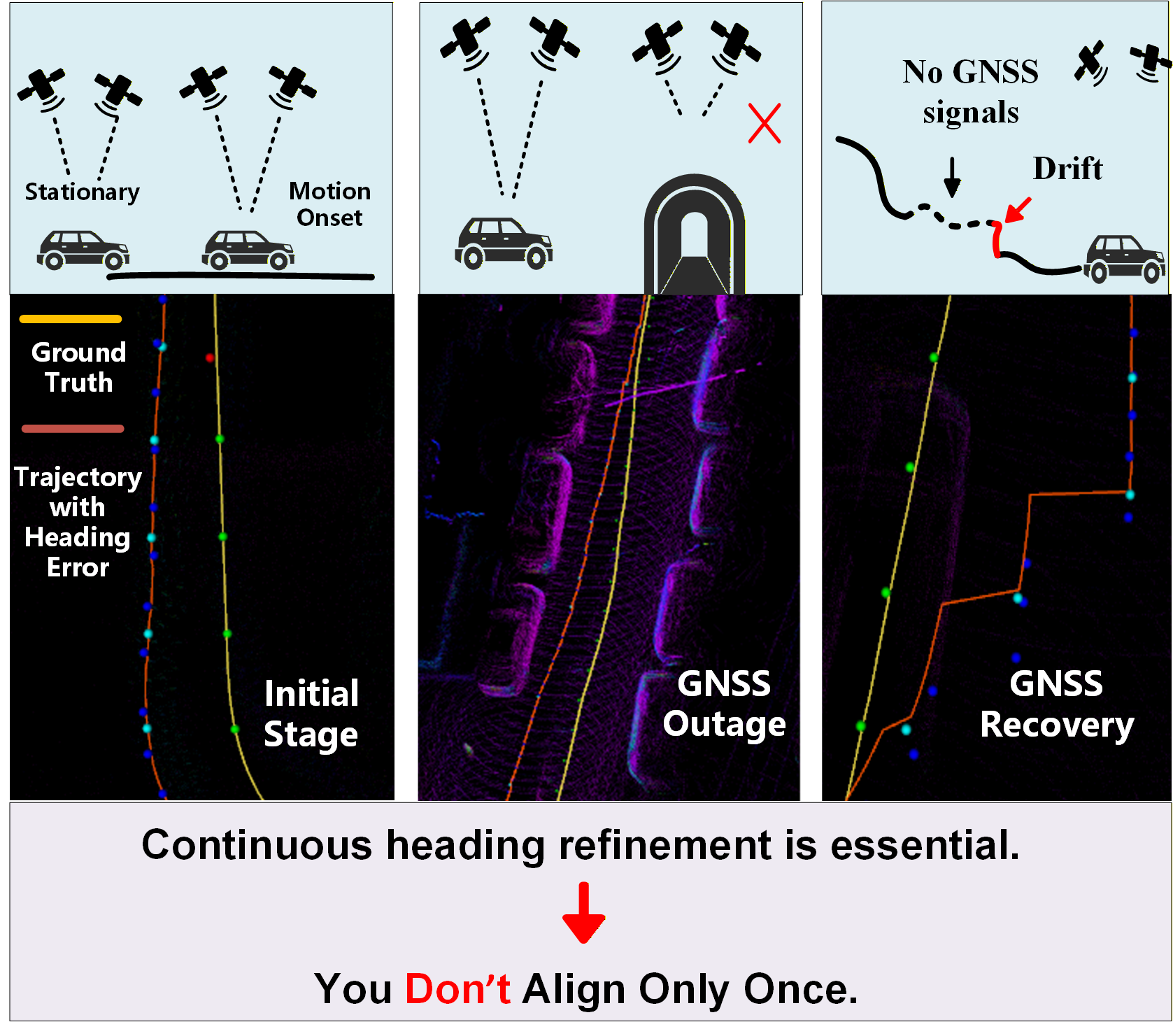}
    \caption{
Impact of inaccurate heading estimation on trajectory consistency. The panels from left to right illustrate cumulative drift originating from initial errors, error propagation during dead-reckoning, and position discontinuity at the point of GNSS recovery. 
    }
    \label{figure1}
\end{figure}

Accurate and reliable pose estimation serves as the foundation for autonomous robots to achieve robust environmental perception and intelligent decision-making. In complex and dynamic scenarios, multi-source fusion systems have emerged as the predominant paradigm, leveraging the complementary strengths of Global Navigation Satellite System (GNSS), Inertial Measurement Unit (IMU), and Light Detection and Ranging (LiDAR) to maintain both global consistency and local precision. Although accelerometers can effectively constrain pitch and roll angles by sensing the gravity vector, heading estimation remains a persistent challenge. Due to the lack of gravitational constraints and the weak observability inherent in stationary or uniform motion, the heading angle is highly susceptible to sensor biases and environmental disturbances. As illustrated in Fig. \ref{figure1}, even marginal heading errors can lead to significant trajectory divergence over time. Consequently, rapid and precise heading estimation represents a critical bottleneck for ensuring spatial alignment and positioning stability, particularly during vehicle startup or prolonged GNSS outages. It is imperative that heading refinement is not treated as a singular initialization step; rather, it must function as a dynamic mechanism capable of being triggered throughout the navigation process to ensure sustained reliability.

These established algorithmic frameworks, while effective in their respective domains, typically focus on initial alignment rather than continuous heading refinement. Specifically, single-epoch instantaneous alignment methods \cite{shao2015study} exhibit a heavy dependence on high-precision velocity vectors or multi-antenna baselines, which may be susceptible to noise during low-speed maneuvers or multipath interference. Recursive filtering methods \cite{huang2017kalman}, while computationally efficient, are often constrained by linearized error models that struggle to suppress rapid drift in IMU biases without continuous external observations. Batch optimization methods \cite{wu2011optimization, kuang2025robust} enhance robustness by incorporating historical data; however, these approaches generally necessitate relatively long, continuous trajectories to ensure convergence. Consequently, many mainstream methodologies are designed for a singular initialization under stable GNSS conditions, potentially lacking the flexibility to adapt to volatile dynamic environments where intermittent heading corrections are required.

To address these challenges, this paper introduces WinTA-GIL, a heading refinement framework that integrates information from GNSS, IMU, and LiDAR. In contrast to conventional methods that rely on continuous GNSS data for one-time alignment, WinTA-GIL utilizes high-precision relative positioning from LiDAR-Inertial Odometry (LIO) to construct short-term temporal windows. Within these windows, local trajectories estimated by the LIO are registered with filtered absolute GNSS observations via non-linear least squares. This formulation transforms heading estimation from incremental dead reckoning into an optimization problem centered on trajectory segment consistency, thereby enabling rapid and repeatable refinement. Furthermore, an adaptive re-estimation mechanism, triggered by state discrimination, automatically corrects the heading during system startup or signal recovery, which effectively suppresses accumulated errors throughout the navigation process.

The main contributions of this paper are summarized as follows:
\begin{itemize}
\item We propose a {GNSS/IMU/LiDAR} heading refinement algorithm based on a temporal window. By employing weighted trajectory matching, heading estimation is shifted from single-point calculation to trajectory-based optimization, significantly improving accuracy.
\item We design an adaptive re-estimation mechanism based on state discrimination that automatically triggers heading correction. This mechanism utilizes observation constraints during the initial recovery phase to correct inertial errors accumulated during outages, enhancing stability in dynamic environments.
\item The effectiveness of the proposed algorithm is validated using both open-source and self-collected datasets. Experimental results demonstrate that WinTA-GIL accurately estimates the heading angle during startup phases or within extremely short signal windows following prolonged {GNSS} outages.
\end{itemize}

\section{Related Work}
\begin{figure}[htbp]
    \centering
    \includegraphics[width=0.47\textwidth]{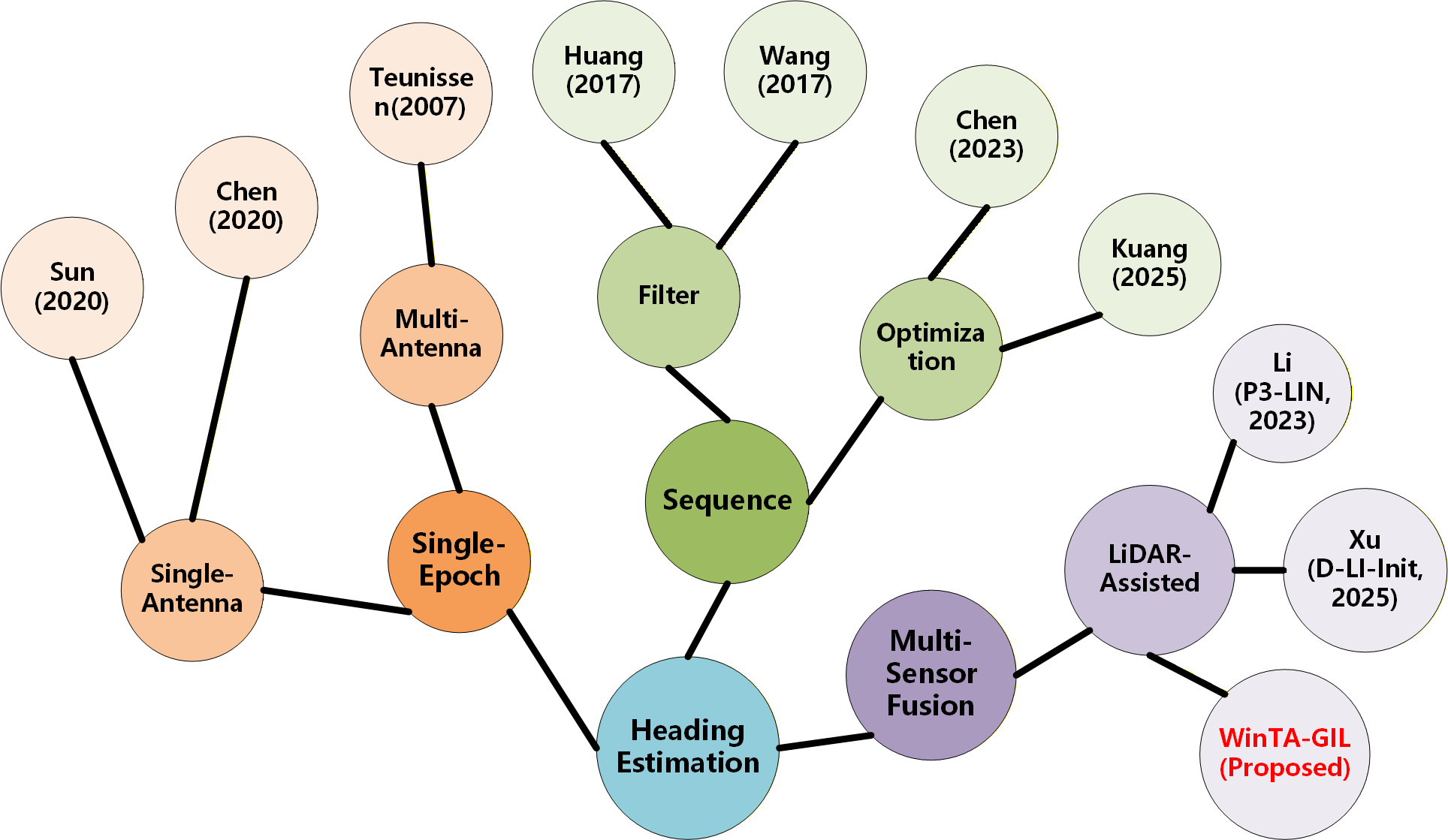}
    \caption{Evolutionary trends and classification of heading estimation.}
    \label{fig:Related work}
\end{figure}

As illustrated in Fig. \ref{fig:Related work}, research in heading alignment is generally categorized into two paradigms: single-epoch based methods and sequence based methods.

\subsection{Single-Epoch Based}
Single-epoch methods determine heading instantaneously using data from a single epoch. High-end IMUs achieve this through gyro-compassing \cite{syed2008civilian}, whereas low-cost systems require external aids. Groves et al. combined a single-axis gyroscope with Global Positioning System (GPS) velocity \cite{groves2009vehicle}, and Shao et al. introduced differential corrections including position, pseudorange, and carrier phase to refine precision \cite{shao2015study}. To mitigate the hardware constraints of single-antenna systems, Sun et al. utilized Time-Differenced Carrier Phase (TDCP) models \cite{sun2020precise} and vehicle kinematics. Furthermore, Chen et al. proposed a rapid calculation based on the similarity of short-term INS/RTK trajectories \cite{chen2020rapid}. Although these methods are effective, they depend either on continuous motion or on multi-antenna baseline vectors \cite{teunissen2007general}, which limits the applicability of such approaches on platforms with low cost or structural constraints.

\subsection{Sequence Based}
In signal-challenged environments, sequence-based methods utilize multi-sensor fusion to enhance robustness, following either filtering or optimization pathways.

Filtering-based methods recursively estimate inertial states using error models. To handle automotive GNSS noise, Shin employed differential corrections \cite{shin2004unscented}, while Han used time-differenced measurements and motion constraints \cite{han2010novel} for gradual convergence. Subsequent frameworks integrated multi-antenna carrier phases \cite{wang2017flight} or gyroscope and velocity constraints \cite{huang2017kalman} to improve the speed of initialization. However, the linearized models in filtering struggle with the strong non-linearity and intermittent observability inherent in urban canyons.

Optimization-based methods minimize residuals over multiple epochs, typically within factor graph frameworks. Wu proposed Optimization-Based Alignment (OBA) \cite{wu2011optimization}, which optimizes the smallest eigenvector of vector observations, later extending it to velocity and position integration \cite{wu2013velocity, wu2014new} for robust alignment under maneuvers. Recent works matched MEMS-DR relative trajectories with GNSS paths \cite{chen2023rapid} and introduced two-step robust strategies to handle outliers \cite{kuang2025robust}.

Despite these advances, GNSS-only methods fail under prolonged outages. While LiDAR-assisted methods such as D-LI-Init \cite{xu2025dynamic} and P3-LINS \cite{li2023p3} offer high-precision constraints, these approaches often require long continuous LiDAR-IMU trajectories and stable GNSS seeds. Achieving reliable and rapid initialization under sparse and intermittent GNSS conditions remains an unresolved challenge that warrants further research.

\section{METHODOLOGY}

\begin{figure*}[t] 
\centering 
\includegraphics[width=0.9\textwidth, height=0.4\textheight, keepaspectratio]{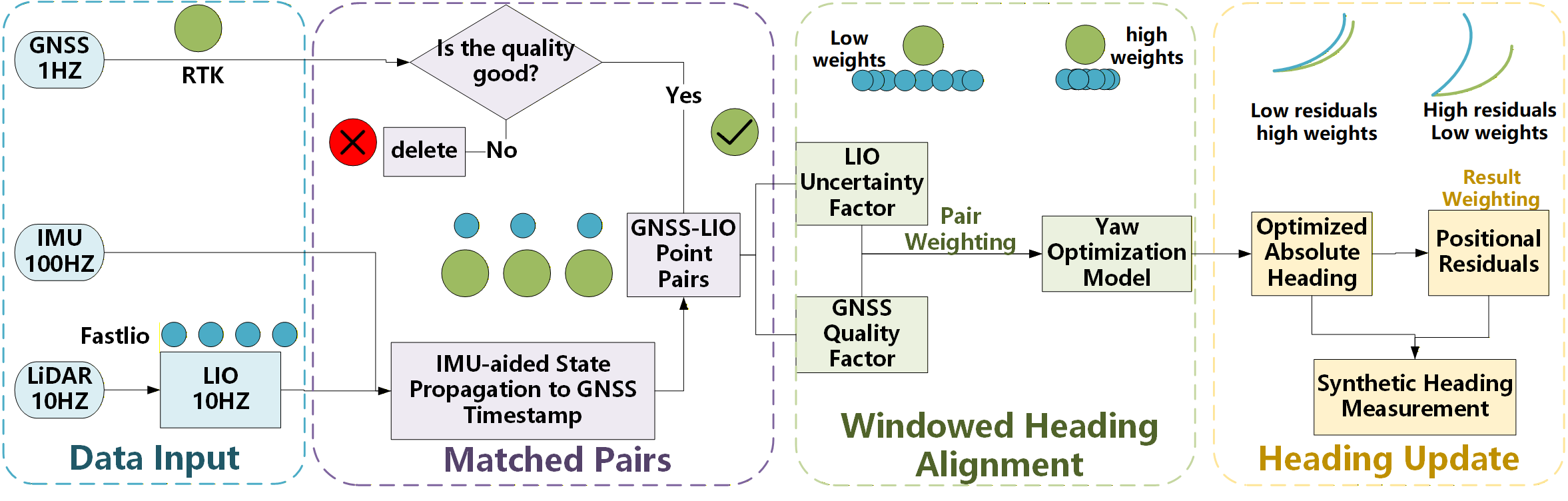} 
\caption{Overview of the proposed WinTA-GIL framework. The system integrates GNSS, IMU, and LIO data through a three-stage refined initialization process: windowed trajectory matching, weighted heading optimization, and pseudo-observation filtering.} 
\label{fig:framework} 
\end{figure*}

To facilitate robust heading estimation, a motion-aware triggering logic is employed to initiate the WinTA-GIL optimization framework, the architecture of which is illustrated in Fig. \ref{fig:framework}.To maintain operational reliability, an adaptive triggering mechanism is first employed as a gatekeeper to monitor both the initial motion onset during the startup phase and the signal-geometry consistency during steady-state GNSS recovery. Once the predefined triggering conditions are satisfied, the WinTA-GIL framework is activated to perform window-based rigid-body registration. This optimization process utilizes a Ceres-based solver and a multidimensional adaptive weighting strategy to align the LIO and GNSS trajectories. By injecting the resulting refined heading as a pseudo-observation into the Extended Kalman Filter (EKF), the system enables the precise correction of the heading angle and sensor biases across diverse dynamic environments.

\subsection{System Overview} 
The input of the system consists of three components: 1Hz GNSS observations, 100Hz IMU data, and 10Hz LiDAR-Inertial Odometry (LIO) outputs. The GNSS trajectory points are obtained via the Real-Time Kinematic(RTK) algorithm \cite{teunissen2015review}, while the LIO trajectory is generated using the Fast-LIO2 framework \cite{xu2022fast}. The proposed refinement initialization framework is divided into three stages: window-based trajectory matching, windowed heading registration, and pseudo-observation injection filtering.

In the window-based trajectory matching stage, a temporal window is established upon detecting vehicle motion. GNSS observation quality within the window is evaluated, and LIO trajectory points are synchronized to the nearest GNSS positioning points. Subsequently, these points are projected into the North-East-Down (NED) navigation frame, resulting in synchronized LIO-GNSS trajectory point pairs. The second stage involves windowed heading registration, where weights are assigned to pairs based on GNSS quality, LIO trajectory uncertainty, and their temporal offset. These weighted data are incorporated into a yaw angle optimization model to determine the refined heading angle. Finally, the reliability of the optimized heading is assessed based on heading variation within the window; this value is injected into the filter as a pseudo-observation for the measurement update.

\subsection{Adaptive Triggering via Motion-Geometry Consistency}

To ensure the reliability of heading refinement and prevent erroneous optimizations caused by GNSS multipath or signal artifacts, a dual-layer state discrimination mechanism is proposed. The WinTA-GIL optimization is triggered only when both the signal precision and the geometric motion patterns are verified.

First, the quality of GNSS observations is evaluated within a temporal sliding window $\mathcal{W}$ of size $N$. For each epoch $k$, a composite standard deviation $\sigma_k$ is defined to characterize the positional uncertainty, which remains consistent with the weighting strategy of the system:
\begin{equation}
\sigma_k = \max(\sigma_{N,k}, \sigma_{E,k}) + 0.5\,\sigma_{D,k}
\end{equation}
where $\sigma_{N,k}$, $\sigma_{E,k}$, and $\sigma_{D,k}$ represent the standard deviations in the North, East, and Down directions at epoch $k$, respectively. This composite metric prioritizes horizontal precision while incorporating vertical uncertainty. The average uncertainty within the window $\mathcal{W}$ is subsequently computed as
\begin{equation}
\bar{\sigma}_{G} = \frac{1}{N} \sum_{k=1}^N \sigma_k
\end{equation}
where $\bar{\sigma}_{G}$ denotes the mean positional uncertainty across the temporal window. The first triggering condition is established as $\bar{\sigma}_{G} < \tau_{\sigma}$, in which $\tau_{\sigma}$ represents the maximum tolerable uncertainty threshold required for a valid alignment.

Second, a Shape Consistency Factor ($\eta$) is introduced based on velocity-profile correlation to verify the geometric fidelity of the motion. The displacement magnitudes for both LiDAR-Inertial Odometry (LIO) and GNSS are extracted as follows:
\begin{equation}
\begin{gathered}
    v_{L,k} = \| \mathbf{p}_{L,k} - \mathbf{p}_{L,k-1} \| \\
    v_{G,k} = \| \mathbf{p}_{G,k} - \mathbf{p}_{G,k-1} \|
\end{gathered}
\end{equation}
where $v_{L,k}$ and $v_{G,k}$ signify the scalar displacements of the LIO and GNSS trajectories at epoch $k$, respectively. The geometric similarity is quantified using the Pearson Correlation Coefficient (PCC) between these two scalar sequences:
\begin{equation}
\eta = \frac{\sum_{k=2}^N (v_{L,k} - \mu_L)(v_{G,k} - \mu_G)}{\sqrt{\sum_{k=2}^N (v_{L,k} - \mu_L)^2} \sqrt{\sum_{k=2}^N (v_{G,k} - \mu_G)^2}}
\end{equation}
where $\mu_{L}$ and $\mu_{G}$ denote the mean displacement magnitudes within the window $\mathcal{W}$ for each source. This correlation coefficient serves to validate whether the GNSS trajectory aligns with the local motion captured by the LiDAR system. Finally, the system activates the alignment process only if the decision function $\mathcal{D}(\mathcal{W})$ is satisfied:
\begin{equation}
\mathcal{D}(\mathcal{W}) = 
\begin{cases} 
1, & \text{if } \eta > \tau_{\eta} \text{ and } \bar{\sigma}_{G} < \tau_{\sigma} \\
0, & \text{otherwise}
\end{cases}
\end{equation}
where $\tau_{\eta}$ is the predefined threshold for shape similarity. This dual-check mechanism ensures that trajectory alignment is performed only when the GNSS observations possess both sufficient precision and geometric congruence with the motion derived from LIO.

\subsection{Matched Pairs}

\begin{figure}[htbp]
    \centering
    \includegraphics[width=0.47\textwidth]{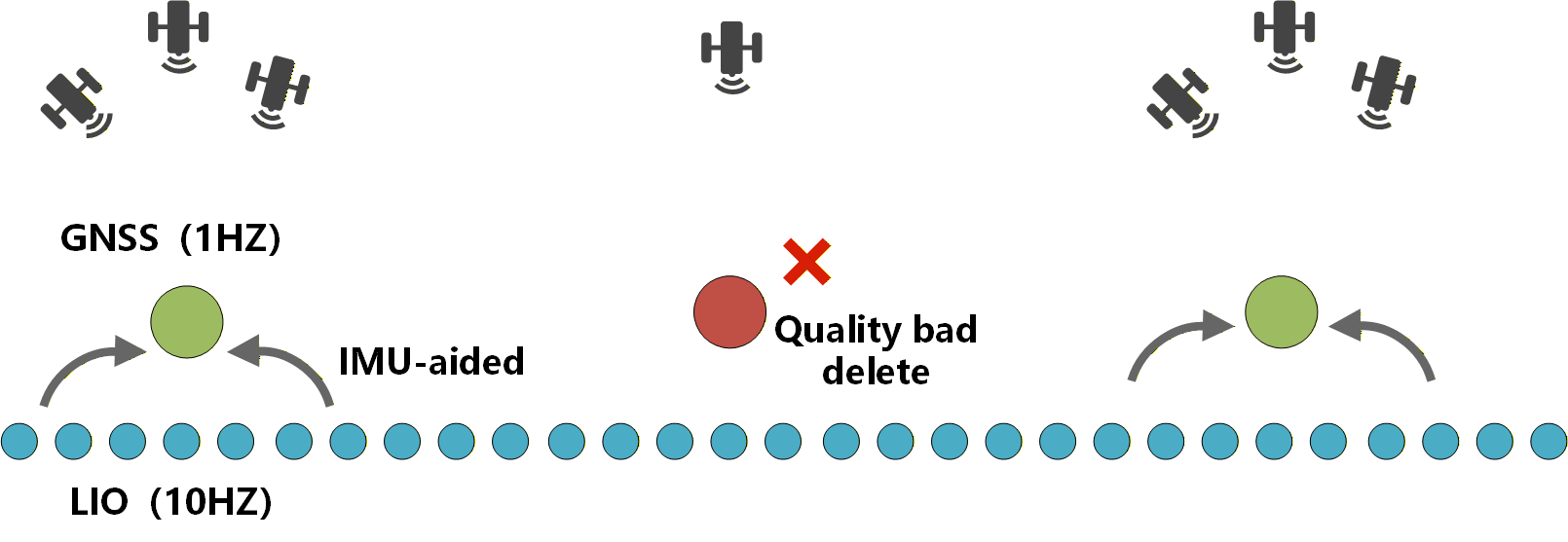}
    \caption{Schematic of the temporal matching and spatial transformation process between GNSS observations and the LIO trajectory.}
    \label{fig:matched pairs}
\end{figure}

WinTA-GIL associates GNSS observations with the LIO trajectory within a predefined temporal window. Each GNSS observation undergoes a rigorous quality assessment, and only those satisfying the established criteria are retained as reliable points. This procedure effectively mitigates the impact of multipath interference and signal degradation in challenging environments. For LIO trajectory points within the window, the system identifies matching GNSS samples based on temporal proximity and utilizes IMU observations to transform the LIO states into the GNSS global coordinate frame. As illustrated in Fig. \ref{fig:matched pairs}, this ensures precise synchronization and spatial alignment in preparation for subsequent registration.

Assuming the LIO sampling time is $t_j$ and the target GNSS observation time is $t_i$, the displacement error primarily originates from the rotational motion occurring within the short interval between these timestamps. By utilizing the IMU angular velocity $\boldsymbol{\omega}$, the rotation increment $\mathbf{R}_{j \to i}$ from $t_j$ to $t_i$ is computed as follows:
\begin{equation}
\mathbf{R}_{j \to i} = \exp\left( \int_{t_j}^{t_i} \left( \boldsymbol{\omega}_t - \mathbf{b}_g \right) dt \right)
\end{equation}
where $\mathbf{b}_g$ represents the gyroscope bias, and the exponential map transforms the integrated angular velocity into a rotation matrix. Subsequently, the LIO position at time $t_j$, denoted as $\mathbf{p}_j^{l}$, is transformed to the synchronized position at time $t_i$:
\begin{equation}
\mathbf{p}_i^{l} = \mathbf{R}_{j \to i} \cdot \mathbf{p}_j^{l}
\end{equation}
where $\mathbf{p}_i^{l}$ serves as the motion-compensated coordinate aligned with the GNSS temporal reference. This transformation effectively eliminates the latency-induced spatial discrepancy between the two independent sensing modalities.

\subsection{Windowed Heading Alignment}

The core of the heading refinement process involves performing rigid-body registration via the Ceres optimization library and extracting an accurate heading angle by minimizing the geometric position error between trajectories. The LIO trajectory is obtained through the Fast-LIO framework and is expressed in the local coordinate frame $l$. 

The primary objective is to design a multidimensional adaptive weighting factor $w_k$ for each pair of matched points $\{ \mathbf{p}_{L,k}^{l}, \mathbf{p}_{G,k}^{n} \}$ to reflect the confidence level of the observations. The cost function is defined as follows:
\begin{equation}
\begin{aligned}
J(\psi, \mathbf{t}) = \sum_{k=1}^{N} w_k \Big\| & \mathbf{p}_{G,k}^{n} - \mathbf{t} \\
& - \mathbf{R}_{l}^{n}(\psi) \left( \mathbf{p}_{L,k}^{l} + \mathbf{R}_{b}^{l} \mathbf{l}^{b} \right) \Big\|^{2}
\end{aligned}
\end{equation}
where $\psi$ and $\mathbf{t}$ denote the heading angle and the translation vector to be optimized, respectively. The weighting factor $w_k$ is assigned to the $k$-th pair based on GNSS observation quality and temporal decay. $\mathbf{p}_{G,k}^{n}$ represents the coordinates of the GNSS positioning point in the NED navigation frame $n$. $\mathbf{R}_{l}^{n}(\psi)$ is the rotation matrix from the local frame $l$ to the navigation frame $n$, parameterized by the heading $\psi$, while $\mathbf{p}_{L,k}^{l}$ denotes the coordinates of the LIO trajectory in the local frame $l$. Furthermore, $\mathbf{R}_{b}^{l}$ is the rotation matrix from the body frame $b$ to the local frame $l$, and $\mathbf{l}^{b}$ represents the lever-arm vector defined in the body frame $b$. 

The weight factor $w_k$ is not a fixed constant; rather, it is constructed as the product of a GNSS quality factor $q_{G,k}$ and an LIO trajectory uncertainty factor $q_{L,k}$:
\begin{equation}
w_k = \max(w_{min}, q_{G,k}) \cdot \max(w_{min}, q_{L,k})
\end{equation}
where $w_{min}$ is a predefined lower bound to prevent numerical instability, and the dual-factor structure ensures that the optimizer prioritizes segments where both sensing modalities exhibit high reliability. The GNSS quality factor $q_{G,k}$ is determined based on the directional standard deviations at epoch $k$:
\begin{equation}
\begin{gathered}
    \sigma_k = \max(\sigma_{N,k}, \sigma_{E,k}) + 0.5\,\sigma_{D,k}, \\
    q_{G,k} = \frac{1}{1 + \sigma_k}
\end{gathered}
\end{equation}
where $\sigma_k$ represents a composite uncertainty metric that accounts for anisotropic noise in the satellite observations. To evaluate the uncertainty of the LIO trajectory during motion compensation, the system computes the weighted mean and covariance of LIO samples. The compactness of the trajectory is characterized by the trace of the covariance matrix, and the corresponding quality factor is defined as:
\begin{equation}
q_{L,k} = \frac{1}{1 + \mathrm{Tr}(\boldsymbol{\Sigma}_k)}
\end{equation}
where $\boldsymbol{\Sigma}_k$ is the covariance matrix of the transformed LIO positions at epoch $k$, and the trace operator $\mathrm{Tr}(\cdot)$ quantifies the total variation of the trajectory distribution. Through this adaptive weighting mechanism, the cost function exhibits robust performance against noise and degraded observations. When GNSS quality deteriorates, the weight $w_k$ is automatically reduced, allowing the optimizer to rely more on high-confidence segments.

\subsection{Heading Update}

After the trajectory alignment and the refinement of the heading $\psi_{\text{after}}$, the system transforms this value into an absolute heading measurement and injects it into the Extended Kalman Filter (EKF). This process utilizes adaptive weighting and state feedback to achieve the precise correction of the navigation state and sensor biases. 

The observation variance $R_\psi = \sigma_\psi^2$ is constructed as follows:
\begin{equation}
\sigma_\psi = \sigma_0 \Delta P
\end{equation}
where $\sigma_0$ is a predefined baseline standard deviation and $\Delta P$ represents the residual position error between the rotated LIO trajectory and the GNSS trajectory.

\section{EXPERIMENT AND ANALYSIS}

\subsection{Experiments Description}

To comprehensively validate the effectiveness of the WinTA-GIL method under diverse data conditions, this study employs a testing scheme that integrates both open-source and self-collected datasets. The experimental evaluation involves five representative data sequences, which are intended to cover a wide range of motion characteristics and GNSS signal conditions. Among these, the Building02 and Street02 sequences are sourced from the open-source i2Nav-Robot\cite{tang2025i2nav} platform and serve primarily as benchmarks for baseline performance. The self-collected datasets include long-distance sequences from a campus environment (Seq.001), short-distance loop sequences under favorable GNSS signal conditions (Seq.002), and a large-scale sequence acquired using an SUV-based platform (Seq.003).

Regarding the hardware configuration, the open-source i2Nav-Robot platform integrates a Livox Mid360 LiDAR, a NovAtel OEM719 receiver, and an ADIS16465 IMU. A high-precision navigation-grade INS is utilized to provide the ground truth, achieving positioning and attitude accuracies of 0.02 meters and 0.01 degrees, respectively. The self-collected experiments utilize both wheeled robots and SUVs as carriers, equipped with a Mid-70 LiDAR, a Honeywell HG4930 IMU, and a NovAtel OEM718D receiver. To ensure the rigor of multi-sensor fusion, all self-collected devices are synchronized based on GNSS timestamps via hardware triggers. Furthermore, both the internal and external parameters of the sensors were precisely calibrated prior to the experiments.

\begin{figure}[htbp]
    \centering
    \includegraphics[width=0.47\textwidth]{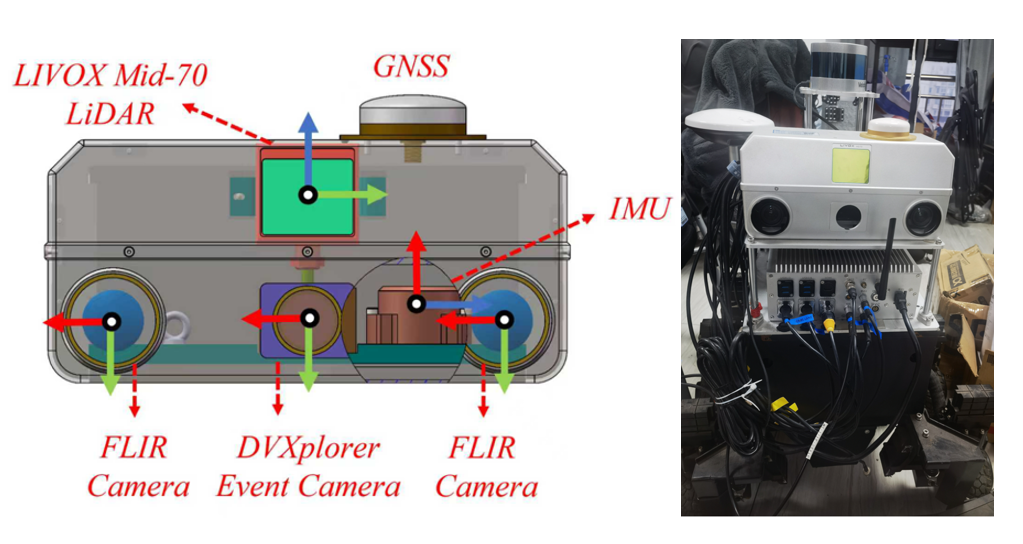}
    \caption{Self-Collected Data Sensor Configuration. \cite{guo2026wecmd}}
    \label{sensors} 
\end{figure}

\begin{table}[ht]
\centering
\caption{Summary of Experimental Datasets}
\label{tab:dataset_overview}
\begin{adjustbox}{valign=c}
\begin{tabular}{cccc}
\toprule
\textbf{Dataset} & \textbf{Source} & \textbf{Platform} & \textbf{GNSS Condition} \\
\midrule
Building02 & Open-source    & Wheeled Robot  & Simulated Outage \\
Street02   & Open-source    & Wheeled Robot  & Simulated Outage \\
Seq. 001   & Self-collected & Wheeled Robot  & Real Outage      \\
Seq. 002   & Self-collected & Wheeled Robot  & Simulated Outage \\
Seq. 003   & Self-collected & SUV Platform   & Real Outage      \\
\bottomrule
\end{tabular}
\end{adjustbox}
\end{table}

In the data processing pipeline, the laser odometry front end of the proposed algorithm adopts the FAST-LIO \cite{xu2022fast} framework, whereas the GNSS-RTK solution is developed using RTKLIB \cite{takasu2009development}. To objectively evaluate the improvements offered by the proposed method, several representative benchmark algorithms are compared, including the pure LiDAR-Inertial Odometry FAST-LIO, the classic GNSS/INS solution KF-GINS \cite{niu2025kf}, the pure GNSS positioning solution RTKLIB, and a baseline GNSS/IMU/LiDAR fusion framework that lacks optimization strategies. The evaluation process focuses on the RMSE for positioning accuracy and attitude precision. A quantitative comparison is performed by considering the convergence speed during initialization and the precision stability at the moment of signal recovery to systematically assess the efficacy of the proposed heading refinement scheme.

\subsection{Performance Evaluation of Initial Heading Refinement}

\begin{figure}[htbp]
\centering
\includegraphics[width=0.45\textwidth]{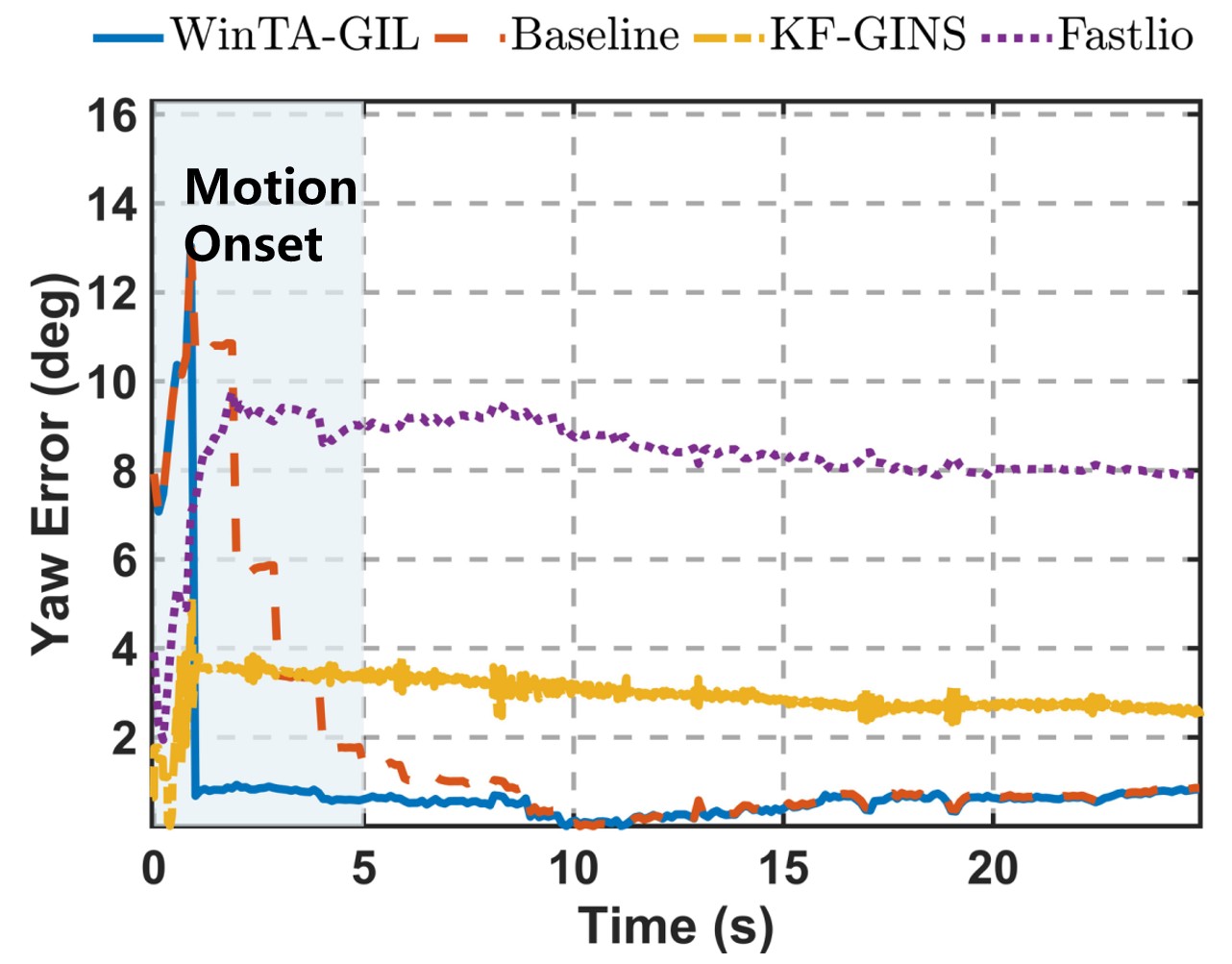}
\caption{Heading estimation error of the Seq. 002 datasets during the first 20 seconds of startup. The proposed WinTA-GIL (blue curve) rapidly reduces the heading estimation error.}
\label{fig:heading_error_startup}
\end{figure}

To evaluate the performance of heading refinement during the initialization stage, experiments were conducted on multiple datasets. As illustrated in Fig. \ref{fig:heading_error_startup}, the blue curve represents the results of the proposed WinTA-GIL method. The experimental results demonstrate that the heading error decreases rapidly to 0.1$^\circ$ within three seconds after the onset of motion. Compared with traditional static initialization and conventional motion-based alignment methods, the proposed approach significantly improves both the accuracy and the convergence speed of heading estimation.

During the low-speed startup phase, the multi-sensor measurements within the sliding window enable rapid and reliable heading estimation, providing a high-precision initial attitude for subsequent global fusion and localization. This rapid heading estimation capability is particularly critical for short-duration tasks, such as those in Seq. 002, as it ensures accurate localization performance during the early stage of robot motion.

\begin{table}[ht]
\centering
\begin{threeparttable}
\caption{Comparison of Heading Estimation Error (RMSE, $^\circ$)}
\label{tab:heading_error}
\small %
\setlength{\tabcolsep}{2pt} 
\begin{tabular*}{\columnwidth}{@{\extracolsep{\fill}}lcccc@{}}
\toprule
\textbf{Dataset} & \textbf{Fast-LIO} & \textbf{KF-GINS} & \textbf{Baseline}$^{\star}$ & \textbf{WinTA-GIL} \\
\midrule
Building02 & 0.58 & 0.51 & 0.68 & \textbf{0.48} \\
Street02   & 1.82 & 2.96 & 1.34 & \textbf{0.79} \\
Seq. 001   & 1.89 & 3.29 & 0.98 & \textbf{0.91} \\
Seq. 002   & 8.09 & 2.65 & 3.11 & \textbf{1.88} \\
Seq. 003   & 1.82 & 1.73 & 1.16 & \textbf{1.13} \\
\midrule
Average    & 2.84 & 2.23 & 1.46 & \textbf{1.04} \\
\bottomrule
\end{tabular*}
\begin{tablenotes}
    \footnotesize
    \item[$\star$] Baseline denotes a fundamental GNSS/IMU/LiDAR fusion framework that lacks optimization strategies.\\ References: Fast-LIO \cite{xu2022fast}, KF-GINS \cite{niu2025kf}.
\end{tablenotes}
\end{threeparttable}
\end{table}

As presented in Table \ref{tab:heading_error}, WinTA-GIL demonstrates a significant improvement in the accuracy of heading estimation during the startup phase across various datasets. The RMSE values for WinTA-GIL are consistently lower than those of the baseline methods, highlighting the effectiveness of the method in rapidly reducing heading error during the initial stage of motion.

\subsection{Heading Refinement Performance under GNSS Signal Recovery in Challenging Environments}

This section evaluates WinTA-GIL under real-world GNSS outage scenarios from Seq.001, focusing on the signal interruption and recovery phases. In traditional frameworks, heading drift accumulated during outages degrades estimation accuracy upon signal restoration. WinTA-GIL addresses this by performing secondary pose estimation within the recovery window.

\begin{figure}[htbp]
    \centering
    \includegraphics[width=0.45\textwidth]{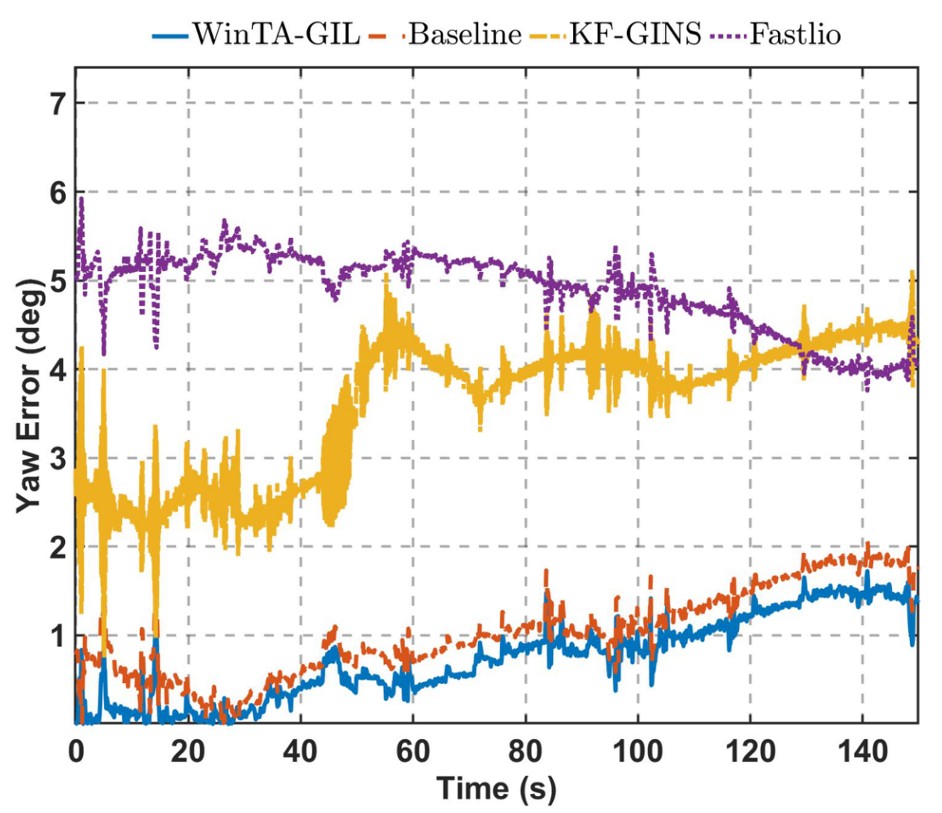}
    \caption{Heading error comparison on Seq.001. WinTA-GIL (blue curve) maintains stable performance, while KF-GINS and Fast-LIO exhibit significant heading errors.}
    \label{001运行过程姿态误差(50S)} 
\end{figure}

\begin{figure}[htbp]
    \centering
    \includegraphics[width=0.45\textwidth]{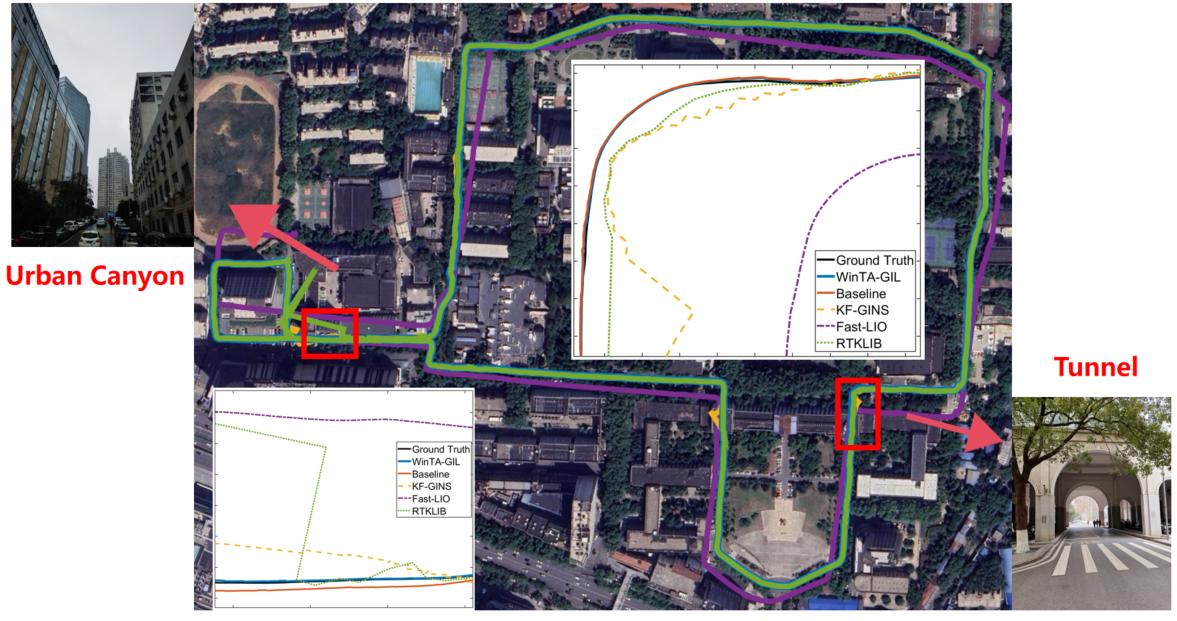}
    \caption{Trajectory comparison on Seq.001 with magnified views of regions with poor GNSS quality. The RTKLIB trajectory exhibits significant drift due to degraded GNSS signals, while WinTA-GIL maintains accurate positioning performance.}
    \label{001轨迹图} 
\end{figure}

Figure \ref{001运行过程姿态误差(50S)} illustrates the attitude errors during operation, while Figure \ref{001轨迹图} presents the positioning trajectories for the Seq.001 dataset. As illustrated in these figures, Fast-LIO (purple curve), serving as a pure LiDAR-inertial odometry, exhibits trajectory drift due to the absence of GNSS corrections. Examination of the magnified regions reveals that the positioning results of RTKLIB (green curve) show significant deviations when GNSS quality is poor. Similarly, KF-GINS, a GNSS/IMU integrated navigation algorithm, suffers from diverging positioning errors as the IMU drifts when GNSS observations become unavailable. In contrast, WinTA-GIL maintains high-precision positioning results even in environments with poor GNSS quality by leveraging LiDAR observations as constraints.

To further evaluate the positioning and attitude accuracy under GNSS recovery scenarios,RMSE metrics were computed for multiple methods. The results are summarized in Table~\ref{tab:recovery_performance}.

\begin{table}[ht]
\centering
\caption{Position and Yaw RMSE on Seq.001}
\label{tab:recovery_performance}
\begin{tabular}{lcc} 
\toprule
\textbf{Method} & \textbf{Position (m)} & \textbf{Yaw (deg)} \\
\midrule
RTKLIB \cite{takasu2009development}   & 1.26 & -- \\
Fast-LIO \cite{xu2022fast}  & 0.19 & 3.09 \\
KF-GINS \cite{niu2025kf}   & 0.18 & 2.49 \\
Baseline  & 0.10 & 0.59 \\
WinTA-GIL & \textbf{0.06} & \textbf{0.25} \\
\bottomrule
\end{tabular}
\end{table}

Experimental results demonstrate that the proposed WinTA-GIL method achieves superior performance in both positioning and attitude estimation. Regarding positioning accuracy, WinTA-GIL yields an RMSE of 0.06\,m, representing a 95.2\% improvement over the pure GNSS solution RTKLIB, and gains of 68.4\% and 66.7\% over Fast-LIO and KF-GINS, respectively. Even compared to the baseline GNSS/IMU/LiDAR fusion framework, WinTA-GIL achieves a 40.0\% enhancement, validating the effectiveness of the heading refinement mechanism.

For attitude estimation, WinTA-GIL shows significant advantages with an RMSE of only 0.25$^\circ$, corresponding to improvements of 91.9\% and 90.0\% over Fast-LIO and KF-GINS. Compared to the 0.59$^\circ$ RMSE of the baseline, the proposed method improves accuracy by 57.6\%. 

Overall, these results validate the superiority and robustness of the WinTA-GIL framework. The significant gains over the baseline confirm that the sliding-window optimization with adaptive heading re-estimation effectively enhances both positioning and attitude accuracy. Specifically, the method provides a reliable heading solution during GNSS signal recovery after prolonged outages, demonstrating strong practicality in real-world autonomous navigation.

\subsection{Ablation Analysis of the WinTA-GIL Framework}

To evaluate the core components of WinTA-GIL, a series of ablation experiments was conducted. By systematically removing or replacing specific modules, we analyzed their contributions to heading estimation accuracy and overall system performance. Five configurations were evaluated. First, WinTA-GIL represents the complete proposed method, incorporating full-window optimization, GNSS quality filtering, IMU-based GNSS-LIO precise matching, and optimized heading observation weights. Second, the Baseline adopts a conventional loosely-coupled GNSS/LiDAR/IMU fusion approach without the proposed window optimization or heading refinement strategies. Third, the No IMU Compensation variant excludes the IMU compensation module, matching GNSS points to LIO trajectory points based solely on the nearest timestamp to verify the necessity of motion compensation. Fourth, the No GNSS Quality Filtering configuration removes the filtering module to assess the system's capability to suppress low-quality observations. Fifth, the Reduced Injection Weight configuration scales the heading observation weight to 50\% of the original value to evaluate the sensitivity of fusion performance to weight parameters.

\begin{figure}[htbp]
\centering
\includegraphics[width=0.45\textwidth]{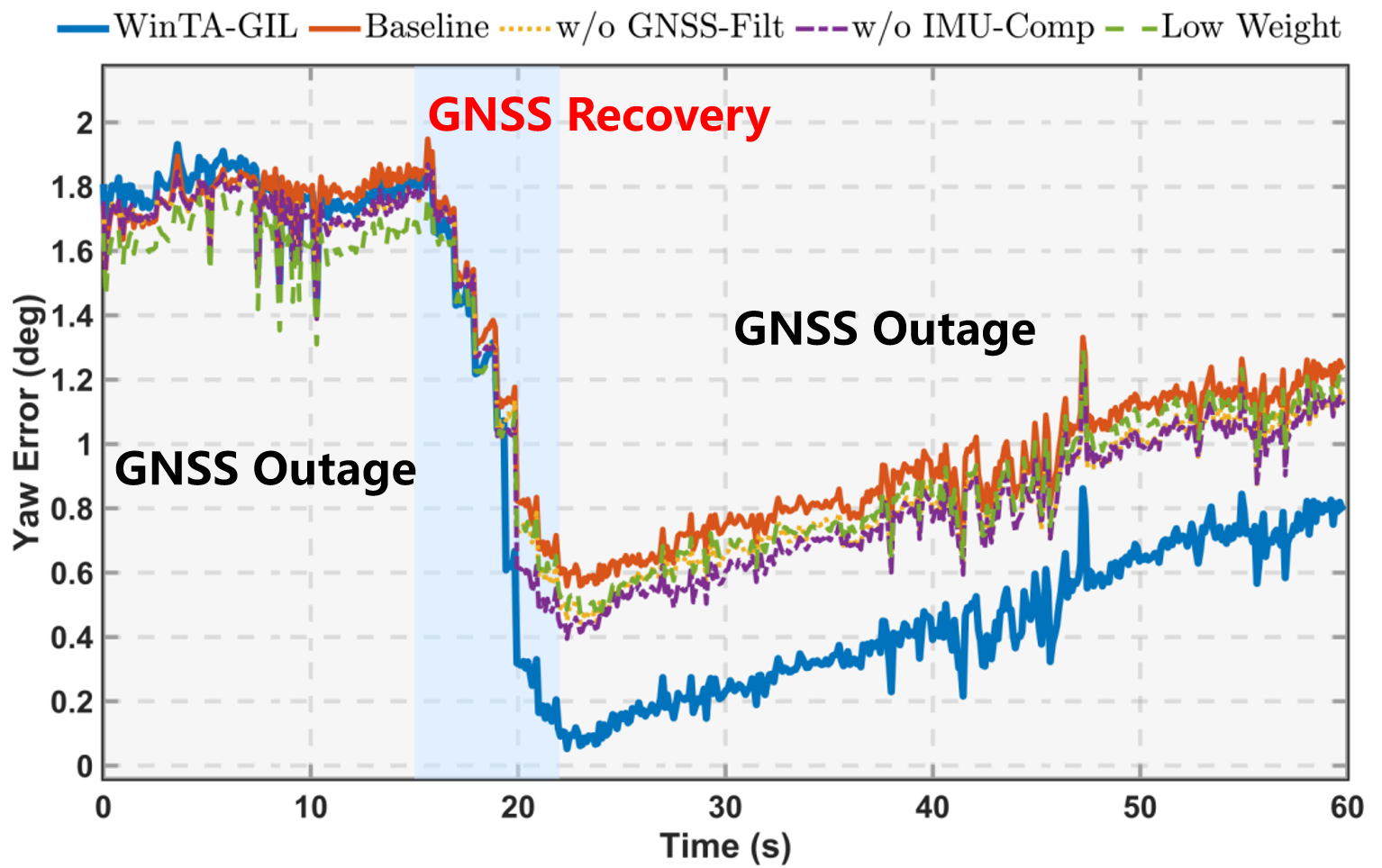}
\caption{Comparison of Heading Estimation Errors Across Ablation Configurations on the Building02 Dataset.}
\label{消融实验航向误差}
\end{figure}

The experiments were conducted on the open-source Building02 dataset, where a 100-second GNSS signal outage was artificially simulated. The evaluation metrics include the heading estimation error ($^\circ$), positioning error (m), and trajectory consistency. Figure \ref{消融实验航向误差} illustrates the comparison of heading estimates during the first 30 seconds of GNSS recovery across different methods.

\begin{figure}[htbp]
\centering
\includegraphics[width=0.4\textwidth]{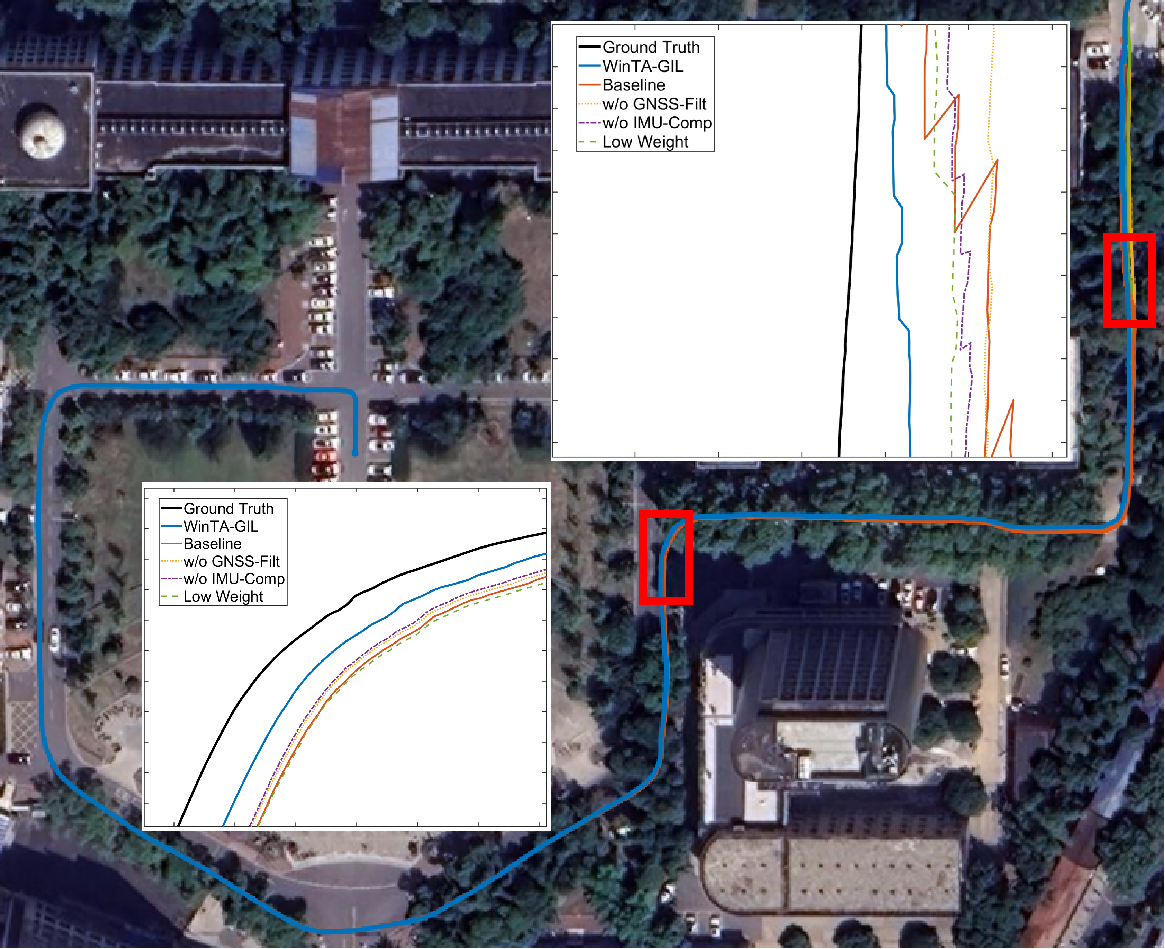}
\caption{Trajectory Comparison of Ablation Configurations on Building02 under Simulated GNSS Denial.}
\label{消融实验轨迹图}
\end{figure}

As illustrated in the figures, the complete WinTA-GIL framework significantly outperforms other variants in the accuracy of heading estimation. Compared to the Baseline, the heading RMSE of WinTA-GIL is reduced by 30\%, validating the effectiveness of the window optimization and heading refinement strategies. The exclusion of GNSS quality filtering results in noticeable heading fluctuations, primarily due to multipath effects and signal occlusion in complex urban environments. Unfiltered low-quality GNSS observations introduce significant errors that degrade the accuracy of the heading estimation. These results emphasize the necessity of the filtering module for outlier rejection and robust estimation. Furthermore, omitting the IMU compensation module increases heading errors because nearest-timestamp matching ignores the motion state differences between GNSS sampling and LIO points. This misalignment is particularly pronounced during high-speed turns or acceleration phases, underscoring the importance of precise temporal synchronization. Finally, reducing the injection weight of pseudo-observations leads to increased errors, as the window-estimated heading provides correct spatial constraints that effectively suppress IMU drift.

Overall, the ablation study confirms the necessity of each module within the WinTA-GIL framework. The IMU compensation ensures temporal consistency, the quality filtering enhances environmental robustness. The synergy of these components enables WinTA-GIL to achieve superior heading estimation and localization performance in challenging environments.

\section{CONCLUSIONS}

This paper presents WinTA-GIL, a heading refinement and re-estimation framework for multi-sensor fusion localization in GNSS-limited environments. To address slow heading convergence during vehicle startup and position jumps from drift after prolonged GNSS outages, we introduce a sliding-window trajectory consistency optimization strategy. This approach enables rapid initialization and dynamic heading correction.

Experimental results across diverse open-source and self-collected multi-platform datasets demonstrate the robustness of WinTA-GIL. The proposed algorithm achieves efficient startup heading alignment and effectively compensates for attitude errors within intermittent GNSS recovery windows, such as those encountered in high-speed highway scenarios. Compared to state-of-the-art (SOTA) methods, WinTA-GIL significantly reduces heading errors—particularly during signal recovery—thereby effectively suppressing position divergence and ensuring continuous, reliable robot localization.

Future work will extend WinTA-GIL to resilient fusion frameworks, providing reliable initial heading for LiDAR reinitialization and enabling high-precision estimation in sparse LiDAR feature or extreme GNSS scenarios.

\addtolength{\textheight}{-12cm}   





\bibliographystyle{IEEEtran}
\bibliography{references}


\end{document}